\documentclass[times]{article}

\usepackage{framed,multirow,multicol}

\usepackage{amssymb}
\usepackage{latexsym}
\usepackage{graphicx}
\usepackage{amsmath}
\usepackage{algorithm}
\usepackage{algorithmic}
\usepackage{bbm}

\usepackage{url}
\usepackage{xcolor}
\definecolor{newcolor}{rgb}{.8,.349,.1}

\def\mat#1{\mathchoice{\mbox{\boldmath $\displaystyle\tt#1$}}
{\mbox{\boldmath$\textstyle\tt#1$}}
{\mbox{\boldmath$\scriptstyle\tt#1$}}
{\mbox{\boldmath$\scriptscriptstyle\tt#1$}}}

\def\vect#1{\mathchoice{\mbox{\boldmath $\displaystyle\bf#1$}}
{\mbox{\boldmath  $\textstyle\bf#1$}}
{\mbox{\boldmath  $\scriptstyle\bf#1$}}
{\mbox{\boldmath  $\scriptscriptstyle\bf#1$}}}

\def\v0{{\vect 0}}

\def\vb{{\vect b}}

\def\vf{{\vect f}}

\def\vh{{\vect h}}

\def\vp{{\vect p}}
\def\vq{{\vect q}}

\def\vr{{\vect r}}

\def\vt{{\vect t}}

\def\vw{{\vect w}}
\def\vx{{\vect x}}
\def\vy{{\vect y}}
\def\vz{{\vect z}}

\def\vD{{\vect D}}

\def\v0{{\vect 0}}

\def\mDelta{{\mat\mDelta}}

\def\mA{{\mat A}}
\def\mB{{\mat B}}

\def\m1{{\mat 1}}

\def\Reales{\mathbb{R}}

\usepackage{authblk}

\title{BEBLID: Boosted Efficient Binary Local Image Descriptor}
\author[1,2]{Iago Su\'arez}
\author[1]{Ghesn Sfeir}
\author[3]{Jos\'e M. Buenaposada}
\author[1]{Luis Baumela}
\affil[1]{Departamento de Inteligencia Artificial. Universidad Polit{\'e}cnica  de Madrid. Campus Montegancedo s/n. 28660 Boadilla del Monte, Spain}
\affil[2]{The Graffter. Centro de Empresas UPM. Campus Montegancedo s/n. 28223 Pozuelo de Alarc{\'o}n, Spain}
\affil[3]{ETSII. Universidad Rey Juan Carlos. C/ Tulip{\'a}n, s/n. 28933 M{\'o}stoles, Spain}
\date{May 2020}                     
\begin{document}

\maketitle

\begin{abstract}
Efficient matching of local image features is a fundamental task in many computer vision applications. However, the real-time performance of top matching algorithms is compromised in computationally limited devices, such as mobile phones or drones, due to the simplicity of their hardware and their finite energy supply.
In this paper we introduce BEBLID, an efficient learned binary image descriptor.
It improves our previous real-valued descriptor, BELID, making it both more efficient for matching and more accurate. To this end we use AdaBoost with an improved weak-learner training scheme that produces better local descriptions. Further, we binarize our descriptor by forcing all weak-learners to have the same weight in the strong learner combination and train it in an unbalanced data set to address the asymmetries arising in matching and retrieval tasks.
In our experiments BEBLID achieves an accuracy close to SIFT and better computational efficiency than ORB, the fastest algorithm in the literature.
\end{abstract}




\section{Introduction}

Local image representations are designed to match images in the presence of strong appearance variations, such as illumination changes or geometric transformations. They are a fundamental component of a wide range of Computer Vision tasks, including 3D reconstruction~\cite{agarwal2009building,schonberger2016structure}, SLAM~\cite{mur2015}, image retrieval~\cite{Nister2006}, tracking~\cite{pernici2014object}, recognition~\cite{lowe1999object} or pose estimation~\cite{wohlhart2015learning}.
They are the most popular image representation approach, because local features are distinctive, view point invariant, robust to partial occlusions and very efficient, since they discard low informative image areas. 

To produce a local image representation we must detect a set of salient image structures and provide a description for each of them. There is a plethora of very efficient detectors for various low level structures such as corners~\cite{rosten2006machine}, segments~\cite{von2010lsd}, lines~\cite{Suarez18} and regions~\cite{matas2002}, that may be described by real valued~\cite{lowe2004distinctive,bay2006surf} or binary~\cite{calonder2010brief,rublee2011orb,alahi2012freak,balntas2015bold,levi2016latch,leutenegger2011brisk} descriptors, being the binary ones the fastest to extract and match. In this paper we address the problem of efficient binary feature description.

Although the SIFT descriptor was introduced twenty years ago~\cite{lowe1999object, lowe2004distinctive}, it is still considered the ``gold standard'' technique. The recent HPatches benchmark has shown, however, that there is still a lot of room for improvement~\cite{balntas2017hpatches}. Modern descriptors based on deep models have boosted the mean Average Precision (mAP) in different tasks~\cite{balntas2017hpatches} at the price of a sharp increase in computational requirements. This prevents their use in hardware and battery limited devices such as smartphones, drones or robots.
This problem has been studied extensively and many local features detectors~\cite{rosten2006machine,von2010lsd,rublee2011orb} and descriptors~\cite{leutenegger2011brisk,calonder2010brief} have emerged, that enable real-time performance on resource limited devices, at the price of an accuracy significantly lower than SIFT.

We have recently introduced BELID~\cite{suarez2019belid}, an efficient real-valued descriptor. Our features use the integral image to efficiently compute the difference between the mean gray values in a pair of square image regions. In BELID we use the BoostedSSC algorithm~\cite{shakhnarovich2005learning} to discriminatively select a set of features and combine them to produce a strong description. BELID achieves execution times close the fastest technique in the literature, ORB~\cite{rublee2011orb}, with an accuracy similar to that of SIFT. Specifically, it provides an accuracy better than SIFT in the patch verification and worse in the image matching and patch retrieval tasks of the HPatches benchmark~\cite{balntas2017hpatches}. Here we use AdaBoost to improve BELID's feature selection procedure and binarize its description.

In this paper we introduce BEBLID (Boosted Efficient Binary Local Image Descriptor), a very efficient binary local image descriptor. We use AdaBoost to train our new descriptor with an unbalanced data set to address the heavily asymmetric image matching problem. To binarize our descriptor we minimize a new similarity loss in which all weak learners share a common weight. In our experiments BEBLID beats both in terms of accuracy and speed  ORB~\cite{rublee2011orb}, the fastest binary descriptor, BinBoost~\cite{trzcinski2015learning} and LATCH~\cite{levi2016latch}, the top performing binary descriptors among the non-Deep Learning literature.


\section{Related work}

SIFT is the most well-known feature detection and description algorithm~\cite{lowe1999object,lowe2004distinctive}. It is widely used because it has a good performance in many Computer Vision tasks. However, it is computationally quite demanding requiring the use of a GPU to achieve real-time performance in certain contexts~\cite{bjorkman2014}.

A number of different descriptors, such as SURF~\cite{bay2006surf}, BRIEF~\cite{calonder2010brief}, BRISK~\cite{leutenegger2011brisk},  ORB~\cite{rublee2011orb}, FREAK~\cite{alahi2012freak}, BOLD~\cite{balntas2015bold} have emerged to speed up SIFT. Binary approaches produce a binary valued descriptor that is very efficient in terms of memory usage and matching speed.
The fastest binary approaches, BRIEF, BRISK, FREAK, ORB, and BOLD, use features based on the comparison of pairs of image pixels. The key for their speed is the use of a limited number of  comparisons selected to be uncorrelated with an unsupervised approach. 
BRIEF uses a fixed size ($9\times9$) smoothing convolution kernel before comparing up to 512 randomly located pixel value pairs. 
BRISK uses a circular pattern, 
smoothing the pixel with a Gaussian of increasing variance the further the pixel is from the center of the pattern.
FREAK chooses uncorrelated pixel pairs from a circular pattern, similar to BRISK, with overlapping Gaussians. 
The ORB descriptor is an extension of BRIEF that takes into account different orientations of the detected local feature. In this case the smoothing is done with an integral image with a fixed sub-window size. It uses a greedy algorithm to uncorrelate the chosen pixel pairs. 
BOLD uses pairwise comparisons estimated like ORB, from which it selects a set of patch adapted comparisons that decrease intra-patch distances. 
The main drawback of these approaches is that they trade accuracy for speed, performing significantly worse than SIFT.

\begin{figure}
	\centering
	\includegraphics[width=1.0\columnwidth]{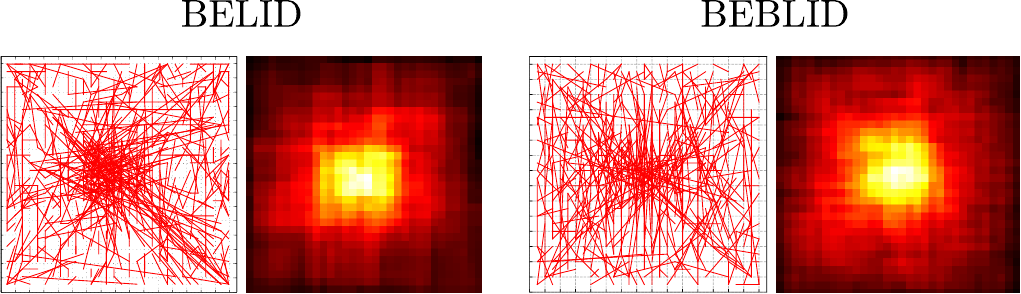}
	\caption{Visualization of BELID and BEBLID pixel location sampling pairs (left) and spatial weight heat maps (right) trained on the Liberty patches data set. Both learn a well distributed set of point pairs giving more importance to the center area.}
	\label{fig:soa-descriptors-patterns}
\end{figure}

Descriptors based on supervised learning algorithms may further improve the performance. 
DAISY~\cite{tola2008fast} learns pooling regions and how to perform dimensionality reduction.
\cite{simonyan2014learning} estimate these hyper-parameters with Convex Optimization, whereas BinBoost~\cite{trzcinski2015learning} and BELID~\cite{suarez2019belid} use Boosting.
The LATCH descriptor~\cite{levi2016latch} compares the gray values in three regions selected to be uncorrelated and discriminative in the patch verification problem.

%
%
Deep Learning enables end-to-end supervised learning of descriptors. CNN-based methods are trained using pairs or triples of cropped patches. Some use Siamese nets~\cite{han2015matchnet}, L2 based loss and hard negative mining~\cite{tian2017l2net} or a modified triplet-based loss~\cite{mishchuk2017working}.
Other methods optimize a loss related to the Average Precision~\cite{he2018local}, an improved triplet loss to help focus on hard examples in training~\cite{wei2018kernelized} or weigh triplets by their difficulty~\cite{zhang2019softmargin}.
L2Net~\cite{tian2017l2net} is the most popular CNN descriptor architecture, which is also used in Hardnet~\cite{mishchuk2017working} and DOAP~\cite{he2018local}.
Few Deep Learning methods address the problem of efficiency in the description. TFeat~\cite{Balntas2016tfeat} uses triplets in a very efficient way for training and a very shallow CNN for speed.
All these methods have improved by a large margin the performance of SIFT in the HPatches benchmark. However, they are computationally more expensive.
TFeat, one of the fastest Deep Learning-based descriptors, running in a GPU is $4\times$ slower than ORB in a CPU~\cite{balntas2017hpatches}. A larger model, such as L2Net, running in a GPU is $15\times$ times slower than ORB in a CPU~\cite{tian2017l2net}.

In this paper we present BEBLID, a binary descriptor that uses a Boosting scheme to select the most discriminative intensity pairwise tests in a local image region (see Fig.~\ref{fig:soa-descriptors-patterns}). 
Like the fastest binary approaches, our features are based on differences of gray values. However, as in BELID, we compute the difference of the mean gray values in a box. The box size represents a scale parameter that improves the discrimination~\cite{suarez2019belid}. In BEBLID, similarly to BinBoost~\cite{trzcinski2015learning}, we search for the best features using a Boosting scheme.
However, each bit in the description produced by BinBoost is a combination of gradient-based features, that are computationally more expensive than simple pairwise tests. In our experiments we prove that our simple and very efficient scaled intensity pairwise tests beat BinBoost's quantized gradient features both in terms of accuracy and speed.


\section{Boosted Efficient Binary Local Image Descriptor}
\label{sec:beblid}

In this section we present our binary image descriptor, BEBLID. To this end, we first introduce a real-valued descriptor based on AdaBoost (see Section~\ref{sec:real-desc}). The use of AdaBoost in our weak learner (WL) selection strategy enables us to train with unbalanced data sets. This is further simplified into a binary descriptor when all WL share the same weights (see Section~\ref{sec:bin-desc}). The key for the efficiency of both descriptors lies in the use of a very efficient WL, based on thresholded pairwise tests computed on square patch regions of arbitrary size (see Section~\ref{sec:average_box_wl}).

\subsection{Real valued Boosted Efficient Local Image Descriptor}
\label{sec:real-desc}

Let $\{(\vx_i, \vy_i, l_i)\}_{i=1}^N$ be a training set composed of pairs of image patches, $\vx_i, \vy_i \in \mathcal{X}$, and labels $l_i \in \{-1,1\}$. Where $l_i = 1$ means that both patches correspond to the same salient image structure and $l_i = -1$ that they are different. 
We use AdaBoost to minimize the loss 
\begin{equation}
\mathcal{L}_{BELID}=\sum_{i=1}^{N} \exp \left(-\gamma l_{i} 
\underbrace{\sum_{k=1}^{K} \alpha_{k} h_{k}\left(\vx_{i}\right) h_{k}\left(\vy_{i}\right)}_{ g_s(\vx_i, \vy_i)}
\right),
\label{eq:loss}
\end{equation}
where $\gamma$ is the shrinkage or learning rate parameter and $h_k(\vz)\equiv h_k(\vz; f, T)$ corresponds to the $k$-th WL combined with weight $\alpha_k$ in the ensemble $g_s$. The WL depends on a feature extraction function $f: \mathcal{X} \rightarrow \mathbb{R}$ and a threshold $T$. Given $f$ and $T$ we define our WL by thresholding $f(\mathbf{x})$ with $T$,
\begin{equation}
h(\vx ; f, T)=\left\{\begin{array}{ll}{+1} & {\text { if } f(\mathbf{x}) \leq T} \\ {-1} & {\text { if } f(\vx)>T}\end{array}\right. .
\label{eq:th-weak-learner-definition}
\end{equation}

\begin{figure}
	\centering
	\includegraphics[width=\columnwidth]{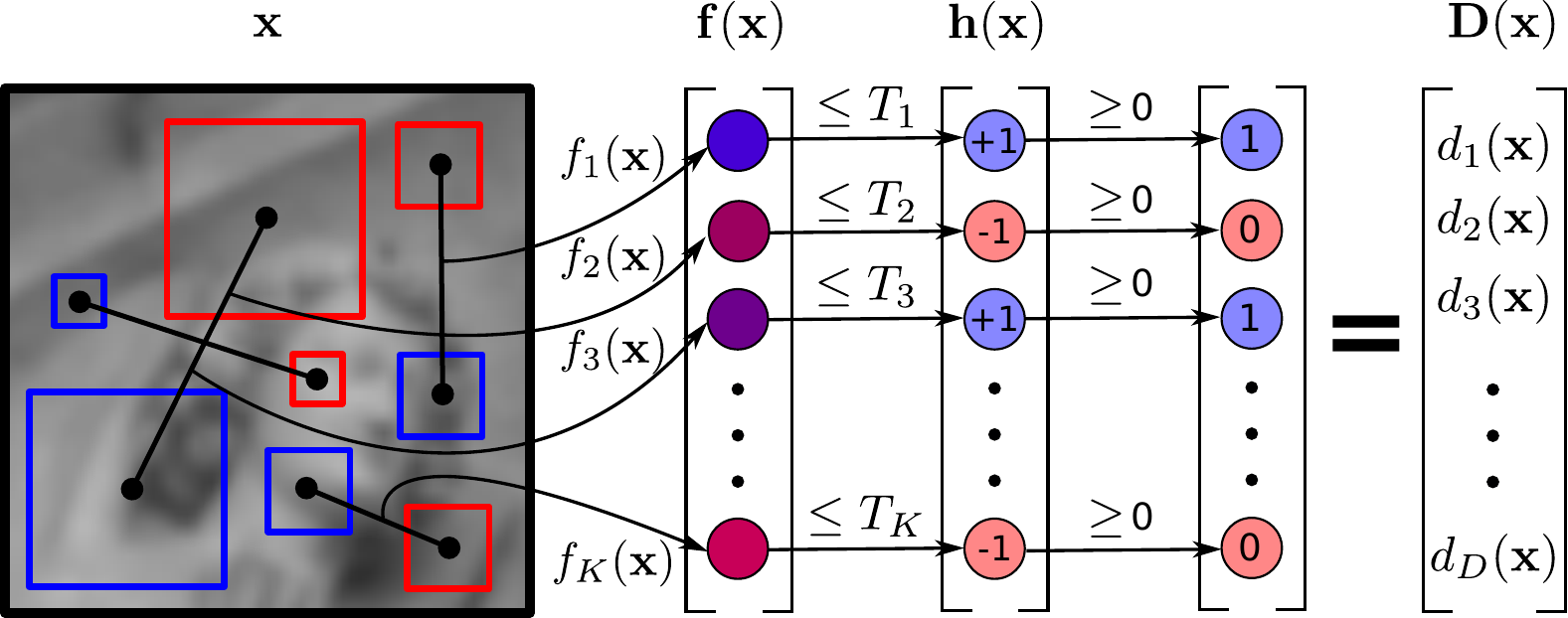}
	\caption{BEBLID descriptor extraction workflow. To describe an image patch, BEBLID efficiently calculates the mean gray value of the pixels in the red and blue boxes. For each pair of red-blue boxes it subtracts their average values obtaining $\vf(\vx)$, the WL. It then thresholds $\vf(\vx)$ to obtain $\vh(\vx)$ and the binary descriptor $\vD(\vx)=\vh(\vx)\geq 0$.}
	\label{fig:beblid-descriptor-diagram}
\end{figure}

The loss in Eq.~\ref{eq:loss} can be seen as a similarity learning function
given by $g_s$ and $\vh(\vx)$ is the vector of $K$ WL responses for image patch $\vx$. The descriptor of this patch is given by 
\begin{equation}
\vD(\vx) = \mA^{\frac{1}{2}}\vh(\vx)=[\sqrt\alpha_1\cdot h_1(\vx), \ldots, \sqrt\alpha_K\cdot h_K(\vx)]^\top
\label{eq:belid_u_descriptor}
\end{equation}
where $\mA=\mbox{diag}(\alpha_1,\ldots, \alpha_k)$ and $\alpha_i$ is the AdaBoost weight for the $i$-th WL, $h_i(\vx)$. We denote this descriptor as BELID-U-ADA (Boosted Efficient Local Image Descriptor, Un-optimized, trained with AdaBoost) in contrast to BELID~\cite{suarez2019belid} that learns a complete matrix $\mA$ modeling the correlations among WLs (see Section~\ref{sec:belid}).

\subsection{Thresholded Average Box weak learner}
\label{sec:average_box_wl}

The key for BEBLID's efficiency is selecting an $f(\vx)$ that is both discriminative and fast to compute.
We define our feature extraction function, $f(\vx)$, 
\begin{equation}
         f(\vx; \vp_1, \vp_2, s) = \frac{1}{s^2}\left(\sum_{\vq \in R(\vp_1, s)} I(\vq)
         - \sum_{\vr \in R(\vp_2, s)} I(\vr)\right),
    \label{eq:ab-wl-eq2}
\end{equation}
where $I(\vt)$ is the gray value at pixel $\vt$ and $R(\vp,s)$ is the square box centered at pixel $\vp$ with size $s$. Thus, $f$ computes the difference between the mean gray values of the pixels in $R(\vp_1, s)$ and $R(\vp_2, s)$. The red and blue squares in Fig.~\ref{fig:beblid-descriptor-diagram} represent, respectively, $R(\vp_2, s)$ and $R(\vp_1, s)$.

On each AdaBoost iteration, we find the best WL by evaluating: 1) a fixed number, $N_p$, of pixel pairs $(p_1, p_2)$; 2) all square regions of size $s$; and 3) all thresholds $T$. 
Inspired by BoostedSSC~\cite{shakhnarovich2005learning} we have developed an efficient algorithm (see Alg.~\ref{alg:threshold-rate}) to select the best discrete threshold for a given WL candidate without an exhaustive evaluation. The algorithm takes as input the responses of $f(\vx; \vp_1, \vp_2, s)$ at each pair of patches and finds the threshold that minimizes the weighted classification error. The algorithm has  $O(P \log P)$ ($P=2N$) complexity that derives from the sorting step in line 9, this allows us for a fast search over all possible thresholds.

\renewcommand{\floatpagefraction}{.8} 
\begin{algorithm}
	\caption{ThresholdRate(P, f, W): Evaluation of projection thresholds given similarity-labeled examples.}
    \label{alg:threshold-rate}
	\textbf{Input:} Set of labeled pairs $P=\{(\vx_i, \vy_i, l_i)\}_{i=1}^{N} \subset \mathcal{X}\times\{-1,1\}$\\
	\textbf{Input:} A feature extraction function $f : \mathcal{X} \rightarrow \mathbb{R}$ \\
	\textbf{Input:} Data weights $W=\left[w_{1}, \ldots, w_{N}\right]$ \\
	\textbf{Output:} $\left\{(T_{t},\mathrm{\epsilon}_{t})\right\}_{t=1}^{n}$,  
	where $\mathrm{\epsilon}_{t}$ is accuracy with threshold $T_{t}.~~~~$\\
	\begin{algorithmic}[1]
		\STATE Let $v_{i, 1} :=f\left(\vx_i\right)$;  $v_{i, 2} :=f\left(\vy_i\right)$ , $i=1, \dots, N$
		\STATE Let $u_{1}<\ldots<u_{n-1}$ be the $n-1$ unique values of $\lfloor v_{i, p}\rceil$
		\STATE Let $\Delta_{j} :=\left(u_{j+1}-u_{j}\right) / 2,$, $j=1, \dots, n-2$
		\STATE Let $T_{1} :=u_{1}-\Delta_{1},$ and $T_{j+1} :=u_{j}+\Delta_{j},$, $j=1, \ldots, n-1$
		\FORALL{$i = 1,...,N$}
			\STATE Let $d_{i, 1} :=\left\{\begin{array}{ll}{-l_i w_i} & {\text { if } v_{i, 1} \leq v_{i, 2}} \\ {+l_iw_i} & {\text { if } v_{i, 1}>v_{i, 2}}\end{array}\right.$
			\STATE Let $d_{i, 2} :=\left\{\begin{array}{ll}{-l_i w_i} & {\text { if } v_{i, 1}>v_{i, 2}} \\ {+l_i w_i} & {\text { if } v_{i, 1} \leq v_{i, 2}}\end{array}\right.$
		\ENDFOR
		\STATE{}
		\COMMENT{ Sort in ascending order by $v_{i, p}$ value: }
		\STATE $\{(v^{(k)}, d^{(k)})\}_{k=1}^{2N} \leftarrow sort\left(\{(v_{i, p}, d_{i, p})\}_{i=1, \ldots, N, p=1,2}\right)$ 
		\STATE $\mathrm{\epsilon}_{0} := \sum_{i = 1}^N \mathbbm{1}\{l_i = +1\} \cdot w_i$ \COMMENT{ $\mathbbm{1}$ is the indicator function }
        \STATE{t:=1}
		\FORALL{$j = 1,...,t$}
			\STATE $\mathrm{\epsilon}_{j} := \mathrm{\epsilon}_{j-1}$
			\WHILE{$v^{(t)} \leq T_{j}$}
			\STATE $\mathrm{\epsilon}_{j} := \mathrm{\epsilon}_{j} + 
			d^{(t)}$
            \STATE{t:=t+1}
			\ENDWHILE
		\ENDFOR
	\end{algorithmic}
\end{algorithm}

To speed up the computation of $f$, we use $S$, the integral of the input image. Once $S$ is available, the sum of gray levels in a square box can be computed with 4 memory accesses and 3 arithmetic operations.
To make our descriptor invariant to euclidean transformations, we orient and scale our measurements with the underlying local structure. 

\subsection{Binary descriptor learning}
\label{sec:bin-desc}

To obtain a binary description we optimize the loss
\begin{equation}
    \mathcal{L}_{BEBLID} = \sum_{i=1}^{N} \exp \left(-\gamma l_i \sum_{k=1}^{K} h_k(\vx) h_k(\vy)\right),
    \label{eq:beblid-loss}
\end{equation}
where $\gamma$ is the common WLs weight. In practice it acts as a shrinkage parameter that determines the training speed. Since we stop the training process if the algorithm is not able to find a WL better than random guessing, $\gamma$ also determines the number of selected WLs. 

Finally, to have a $\{0,1\}$ output, we convert the -1 output to 0 and the +1 output to 1 (see Fig.~\ref{fig:beblid-descriptor-diagram}). This new binary descriptor is termed BEBLID, that stands for Boosted Efficient Binary Local Image Descriptor.

This is a Boosting scheme in which all WLs have the same contribution to the final strong decision. The intuition behind this scheme is the following. 
In an AdaBoost-based minimization, such as that used to obtain the BELID-U-ADA descriptor in Section~\ref{sec:real-desc}, the contribution of each WL is weighted by $\alpha_k$. This constant depends on the success of the k-th WL in solving a binary patch classification problem. However, we are interested on using our descriptor for solving many other image related problems. 
So, the $\alpha_k$s  are biased by the patch verification problem used to compute them. A descriptor in which all WLs have the same weight actually performs better in other tasks, such as for example image matching and retrieval. In our experiments we prove that this intuition is correct.

\subsection{BELID, BELID-U and BELID-U-ADA}
\label{sec:belid}

In our previous work~\cite{suarez2019belid} we used BoostedSSC~\cite{shakhnarovich2005learning} to compute the BELID-U descriptor by minimizing Eq.~\ref{eq:loss}.
As described in Section~\ref{sec:real-desc}, in this paper we also optimize it with AdaBoost to produce BELID-U-ADA.

Further, estimating the whole matrix $\mA$ improves the similarity by modeling the correlation between WLs. 
FP-Boost~\cite{trzcinski2015learning} estimates a symmetric $\mA$ minimizing
\begin{equation}
\mathcal{L}_{F P}=\sum_{i=1}^{N} \exp \left(-l_{i} \vh(\vx)^\top \mA \vh(\vy) \right)
\label{eq:fp-boost-loss}
\end{equation}
with Stochastic Gradient Descent.

BELID (Boosted Efficient Local Image Descriptor)~\cite{suarez2019belid} describes an image patch $\vx$ as $\vD(\vx) = \mB^\top \vh(\vx)$, where $\mB=\left[\vb_{1},\cdots,\vb_{D}\right], \vb \in \Reales^{K}$ are the eigenvectors associated to the $D$ largest eigenvalues of matrix $\mA$, estimated with FP-Boost.


\section{Experiments}
\label{sec:experiments}

In our experiments we train our models with the popular Brown data set~\cite{winder2007learning}\footnote{\url{http://matthewalunbrown.com/patchdata/patchdata.html}}. It contains $64\times64$ SIFT detected and cropped gray level image patches from three different scenes: Notre Dame cathedral, Yosemite National Park and Liberty statue in New York.

We evaluate our results with the HPatches benchmark~\cite{balntas2017hpatches}.
It provides patches extracted from images of various scenes under different capturing conditions and tested in verification, matching and retrieval tasks. The set of images are organized in 6 splits: ``a'', ``b'', ``c'', ``illum'', ``view'', and ``full''. In the experiments of this paper we use the ``full'' split that contains all the scenes in the dataset, whereas in~\cite{suarez2019belid} we tested our models in the ``a'' split.

We evaluate the performance using three measures:
\begin{itemize}
    \item \textbf{FPR-95}. False Positive Rate at 95\% recall in a patch verification problem. 
    \item \textbf{AUC}. Area Under the ROC Curve in a patch verification problem. It is a good global measure since it considers all the curve operation points. 
    \item \textbf{mAP}. Mean Average Precision, as defined in the HPatches benchmark, for each of the three tasks: patch verification, image matching and patch retrieval.
\end{itemize}

We have implemented AdaBoost and the learning and testing part of the Thresholded Average Box WL in Python.  Using OpenCV 4.1.0 we have also implemented a C++ version\footnote{The C++ code with the pre-trained descriptors BEBLID-256-M and BEBLID-512-M ( explained in section \ref{sec:exp4-soa-comparison} ) has been made public in \url{https://github.com/iago-suarez/BEBLID}} of our descriptor extraction algorithms.
We use this implementation to evaluate their execution time in Section~\ref{sec:exp-times}.

In all our experiments we train our models with the Liberty Statue patches scaled to $32\times32$ pixels. The size values in the Average Box WL are $\mathcal{S}=\{3, 5, 7, 9, 11, 13, 15\}$, and its location constrained to fall inside the image. We quantize $f(\vx)$ to an integer, to reduce the set $\mathcal{T}$ of WL thresholds. 

\subsection{AdaBoost vs. BoostedSSC}
\label{sec:adaboost_comparison}

In the first experiment we compare AdaBoost with BoostedSSC in verification and evaluate the relevance of selecting a good WL.
First we train our model with BoostedSSC, a BELID-U descriptor as in~\cite{suarez2019belid}. Then we train three versions of the BELID-U-ADA descriptor:
\begin{itemize}
    \item \textbf{BELID-U-ADA-Rand}. In each AdaBoost iteration we use 500 candidate WLs randomly selecting location $(p_1, p_2)$, 
    scale $s$ from $\mathcal{S}$ and threshold $T$ from $\mathcal{T}$.
    \item \textbf{BELID-U-ADA}. In each AdaBoost iteration we randomly select $N_p=500$ candidate pixel locations $(p_1, p_2)$. Then we exhaustively evaluate all scales $s\in\mathcal{S}$ and all thresholds $t\in\mathcal{T}$ for each $(p_1, p_2)$ pair. 
    \item \textbf{BELID-U-ADA-Balanced}. Same as in BELID-U-ADA, but in this case we normalize the data weights to sum 0.5 for the negative and positive classes. 
\end{itemize}

\begin{figure}
	\centering
	\includegraphics[width=0.7\columnwidth]{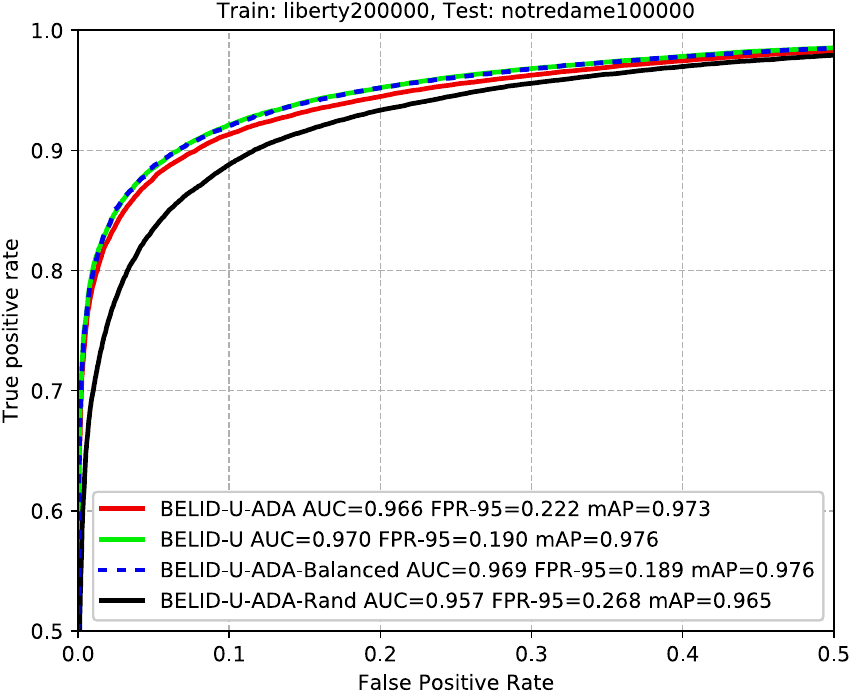}
	\caption{ROC Curve for the verification task in the Brown data sets. We compare BoostedSSC with AdaBoost selecting in each iteration the best WL of a random selection (BELID-U-ADA-Rand), or by exhaustively searching for some of the WL parameters (BELID-U-ADA), or further normalizing the weights of the positive and negative classes. 
	}
	\label{fig:ada-ablation-study}
\end{figure}

In Fig.~\ref{fig:ada-ablation-study} we show the results for the descriptors trained on a balanced data set of 200K patches pairs from the Liberty scene. We test the methods in a balanced data set of 100K patches pairs from the Notredame scene. 
We observe a reduction in FPR-95 from 26.8\%, with BELID-U-ADA-Rand, to 22.2\%, with BELID-U-ADA. Since the only difference between both algorithms is the exhaustive search along the scale, $s$, and threshold, $T$, parameters in the BELID-U-ADA approach, then we can infer 
that to achieve top performance it is important to search for good WLs. This justifies the use of Alg.~\ref{alg:threshold-rate} to speed up the optimal threshold search.  Also, the performance of BELID-U-ADA-Balanced is equivalent to that of BELID-U, based on BoostedSSC. Hence we experimentally  prove that BoostedSSC is just AdaBoost with the assumption of equal priors for positive and negative classes. Finally, an equal prior algorithm, BELID-U-ADA-Balanced, marginally improves the performance of BELID-U-ADA in this balanced verification problem. This is a first hint of the importance of selecting the appropriate priors when training the descriptor. In the next section we analyze this in more detail.

\subsection{Asymmetric training}
\label{sec:asymmetric_problems}
 
Here we exploit the fact that our key problems, matching and retrieval, are asymmetric. We evaluate the performance of descriptors trained with AdaBoost using unbalanced data sets from the Liberty scene.
In our experiments we fix the number of training data to 1 million. 
To train with a data set with X\% positives, we first randomly select X*10k positive samples. Then randomly generate negative pairs up to 1 million. 
In Table~\ref{tab:best-beblid-unbalance-ratio} we show the results for BELID-U-ADA and BEBLID descriptors with 512 components. In Table~\ref{tab:learning-rates} we provide the learning rates used when training these descriptors.

\begin{table*}
\centering
\caption{Results in the ``full'' split of HPatches when training with different ratios of positive samples (50\%, 20\% and 5\%) from the Liberty data set.BEBLID-U-ADA uses 512 floatin point components and BEBLID uses 512 bits}
\label{tab:best-beblid-unbalance-ratio}
\footnotesize
\setlength{\tabcolsep}{1pt}
\begin{tabular}{l|ccc|ccc|ccc|ccc}
& \multicolumn{3}{c}{Verific. - balanced (AUC)} & \multicolumn{3}{c}{Verification (mAP)} & \multicolumn{3}{c}{Matching (mAP)} & \multicolumn{3}{c}{Retrieval (mAP)} \\
& 50\%    & 20\%    & 5\%  & 50\%    & 20\%    & 5\%     & 50\%    & 20\%    &   5\%   & 50\%    & 20\%    & 5\%     \\ \hline
BELID-U-ADA & 85.59 & 85.41 & \underline{84.33\%} & 67.41\% & 67.34\% & \underline{66.17\%} & 17.89\% & 18.73\% & \underline{20.11\%} & 30.00\% & 30.79\% & \underline{32.22\%} \\ \hline
BEBLID &  85.52 & 84.97 & 84.44 & 67.14\% & 67.31\% & 66.52\% & 17.44\% & 21.84\% & 21.69\% & 29.87\% & 33.82\% & 33.74\% \\ \hline
\end{tabular}
\end{table*}

The test set for the HPatches verification problem consists of 200K positives and 1 million negative examples (see "Verification" results in Table~\ref{tab:best-beblid-unbalance-ratio}). However, we have also added results for a fully balanced test set (see "Verification-balanced" in Table~\ref{tab:best-beblid-unbalance-ratio}).

\begin{table}
\centering
\caption{Learning rates for the descriptors in Table~\ref{tab:best-beblid-unbalance-ratio}.}
\label{tab:learning-rates}
\footnotesize
\begin{tabular}{l|ccc}
& 50\%    & 20\%    & 5\%  \\ \hline
BELID-U-ADA (512f) & 0.1 & 0.1 & 0.4 \\ \hline
BEBLID (512b) & 0.0055 & 0.0055 & 0.0025 \\ \hline
\end{tabular}
\end{table}

The BELID-U-ADA results in Table~\ref{tab:best-beblid-unbalance-ratio} for the verification-balanced problem provide the best AUC=85.59\% with the balanced training data set. We get the best result for matching,  mAP=20.11\%, with the most unbalanced set (5\% positives). In the retrieval problem we also get the best results with the most unbalanced training set. With BEBLID we have similar results. We get the best descriptor for the verification-balanced with the balanced training set, AUC=85.52\%.  
In matching and retrieval we also get the best result with an unbalanced training.
However, in this case, the results with 5\% positives are worse than those with 20\%. We speculate this is due to the small number of positives, 50K pairs in this case, that is scarce for training a BEBLID descriptor. We have experienced the same problem 
with 1\% positives.

In summary, dealing with the asymmetry in the target problem is fundamental to improve the accuracy of image descriptors. Specifically, matching and retrieval tasks on one side, and verification on the other, need different descriptors. Here we have considered the use of AdaBoost trained with unbalanced data sets to address this issue.

\subsection{Tuning BEBLID learning rate}

\begin{figure}
	\centering
	\includegraphics[width=0.8\columnwidth]{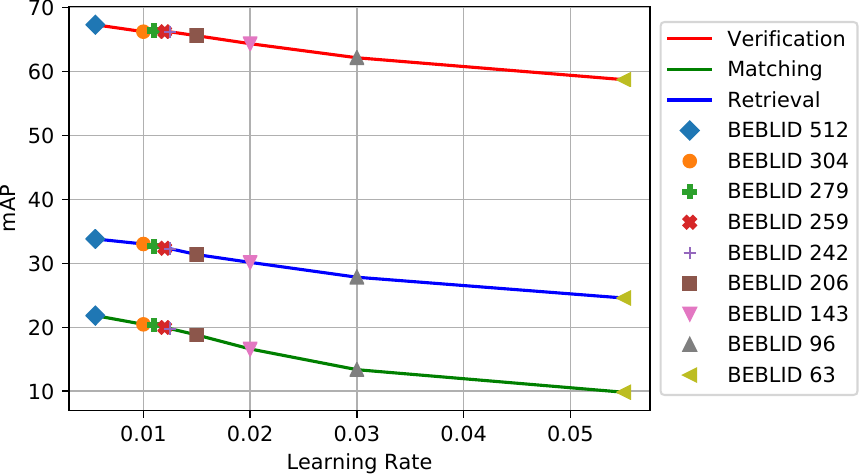}
	\caption{BEBLID learning rate selection experiment. We show the mAP for verification, matching and retrieval in the ``full'' split of Hpatches for models trained with different learning rates, $\gamma$, in the Liberty data set.}
	\label{fig:best-beblid-rl}
\end{figure}

In this experiment we use a training set from Liberty with 20\% positives, selected as described in section \ref{sec:asymmetric_problems}.
We train BEBLID with different learning rates, $\gamma$. In Fig.~\ref{fig:best-beblid-rl} we show the accuracy results in HPatches. We also display the number of bits of the resultant descriptor. 
As expected, the larger the learning rate the lower the number of iterations and bits of the descriptor. To get the desired number of bits (=WLs), $K$, we select a small enough value for $\gamma$ and keep the first $K$ WLs, that are also the most significant ones. We get the best results with 512 WLs and $\gamma=0.0055$.

\subsection{Comparison with the state-of-the-art}
\label{sec:exp4-soa-comparison}

\begin{figure*}
In this case we use the implementation of BinBoost in OpenCV, BinBoost$_{32}$-256, with a descriptor of 256 bits and 32 gradient based WLs per bit, with 8192 WLs evaluated per descriptor. 
\includegraphics[width=1.0\textwidth]{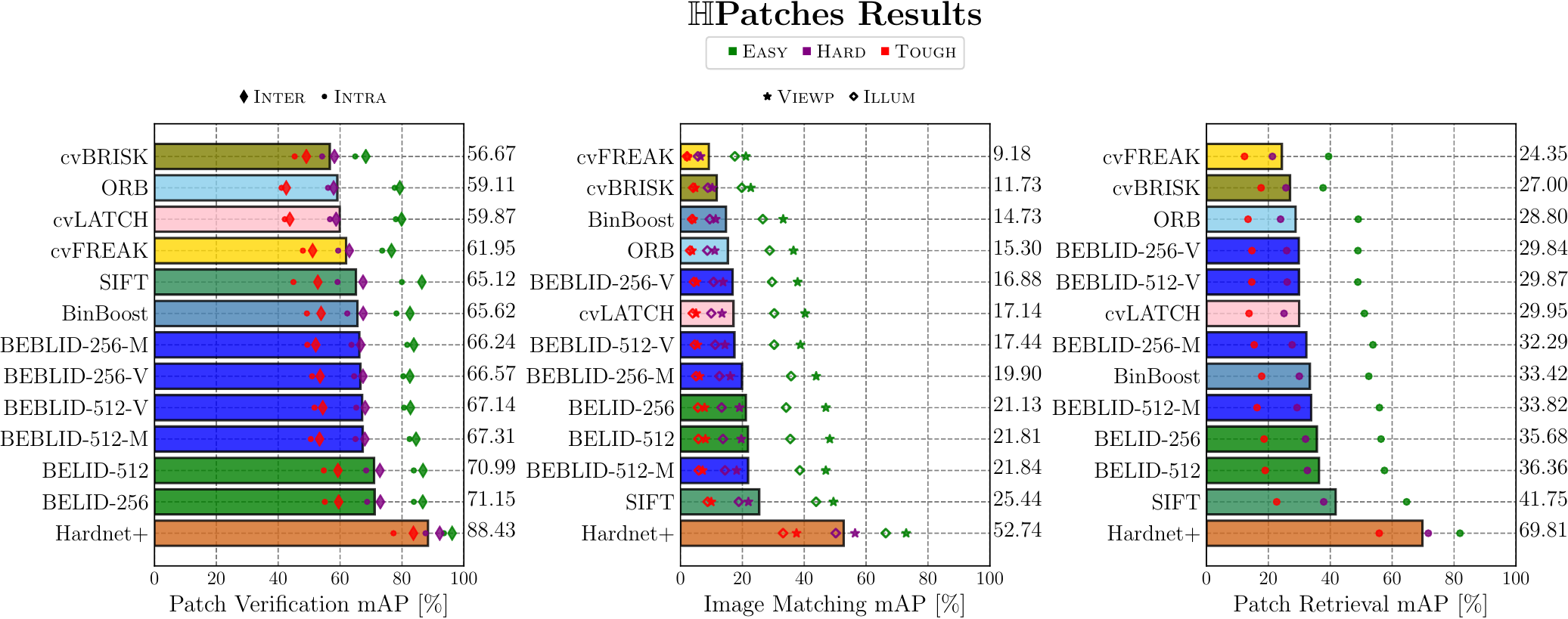}
\caption{Comparison of the state-of-the-art descriptors in the ``full'' split of the HPatches data set. The marker color indicates the noise level: EASY, HARD, and TOUGH. INTRA makers in Patch Verification task corresponds to Patch pairs obtained from the same sequence whereas INTER markers are from different ones. In the Image Matching task, the VIEWP markers refer to scenes with viewpoint distortions and the ILLUM to scenes with illumination changes. The bar length represents the mean of the six variants of each task.} 
\label{fig:hpatches-results}
\end{figure*}

In this section we compare our binary descriptor with the most relevant approaches in the literature.
Fig.~\ref{fig:hpatches-results} shows the results of various BEBLID configurations and those of other competitors. 
Here we compare our binary descriptor trained for the balanced verification problem (``V'' suffix) and for the matching problem (``M'' suffix) with ORB, BRISK, FREAK, LATCH, BinBoost, BELID and SIFT. We train descriptors with suffixes ``M'' and ``V'' with 20\% and 50\% positives respectively (see section~\ref{sec:asymmetric_problems}). We do not display results for BOLD and BRIEF since they are respectively worse than BinBoost and ORB~\cite{balntas2017hpatches}.
We use the OpenCV implementation of BRISK, FREAK and LATCH. The results of SIFT, ORB and BinBoost come from the HPatches benchmark database. 

In the HPatches verification testing set, with 16.66\% positives (200K positives, 1M negatives), all boosting-based descriptors (BELID, BEBLID, BinBoost) are better than SIFT (mAP=65.12\%) while LATCH, ORB, FREAK are worse. The real-valued BELID descriptors~\cite{suarez2019belid}, trained in the balanced Liberty data set, get the best results among non-CNN descriptors. The performance of BEBLID is behind that of BELID because of the binarization and because it does not take into account the correlations between WLs. The balanced version of our new  binary descriptor, BEBLID-512-V, is marginally behind BEBLID-512-M because of the unbalanced testing set. Moreover, BEBLID-512-M, with mAP=67.31\%, and all other versions of BEBLID are better than BinBoost, the best binary descriptor, and SIFT. This is remarkable since both use gradient based features whereas BEBLID uses simple average gray level differences. This is not surprising, however,  since gray level differences for different box sizes is an approximation to the gradient at different scales.

Our best binary descriptor in the matching problem is BEBLID-512-M. Trained with unbalance data it gets mAP=21.84\%, which is worse than SIFT, mAP=25.44\% but, as we will see in the next section, it is two orders of magnitude faster.
BRISK and FREAK are the worse descriptors both in image matching and patch retrieval.
In the former problem BinBoost also shows poor accuracy, mAP=14.73\%. Here the main difference with the patch verification problem is the asymmetry of the matching problem. Two key differences between BEBLID-512-M and BinBoost are that we use unbalanced training and a simpler 
ensemble with common weights.
LATCH gets better results in matching than FREAK, BRISK, BinBoost and ORB. However, in patch retrieval, it is worse than BinBoost. 
BEBLID-512-M beats all its binary competitors both in the image matching and patch retrieval problems.
This result validates our decisions in Section~\ref{sec:bin-desc}.

The number of bits used by a binary descriptor is important. When we halve the number of bits, from 512 to 256, the performance of BEBLID-M drops 1.07 in verification, 1.94 in matching and 1.53 in retrieval. Something similar happens with BEBLID-V. Here BEBLID-256-M and BEBLID-256-V are comparable with ORB and BinBoost, since all of them use 256 bits. In this case we beat both descriptors in matching. In patch retrieval, BEBLID-256-M is marginally worse than BinBoost. However, we get marginally better results than BinBoost using 512 simple WLs (BEBLID-512-M) while BinBoost uses 
gradient based WLs. In the next section we will see that this is an important drawback for BinBoost in terms of efficiency.

We have added to Fig.~\ref{fig:hpatches-results} Hardnet~\cite{mishchuk2017working}, a representative CNN-based descriptor. Hardnet beats by a large margin all handcrafted and learned descriptors, but it has much higher computational requirements.

In summary, we have shown that with our approach we get the best accuracy among non CNN-based binary descriptors in the verification, matching and retrieval problems. This is due to the two key ideas: WLs based on thresholded and scaled pairwise comparisons, and the adaptation of the training process to the level of asymmetry of the problem. 

\subsection{Execution time in different platforms}
\label{sec:exp-times}

In the last experiment we test the C++ implementation of BEBLID processing full images (not cropped patches) in a desktop CPU, Intel Core i7-8750H, and in two limited CPUs, Exynox Octa 7870 and Snapdragon 855. We report the description execution time in the \cite{mikolajczyk2005performance} data set, composed by 48 $800\times640$ images from 8 different scenes. In each of them we detect a maximum of 2000 local structures with SURF.
In this case we use the implementation of BinBoost in OpenCV, BinBoost$_{32}$-256, with a descriptor of 256 bits and 32 gradient based WLs per bit, with 8192 WLs evaluated per descriptor. 

In Table~\ref{tab:execution_times} we show the average execution time per image and the size of each  descriptor in terms of the number of components, that can be floating point numbers (f) or bits (b).
We compare the execution time with C++ implementations of BELID~\cite{suarez2019belid} and other relevant descriptors in the OpenCV library: SIFT~\cite{lowe2004distinctive}, ORB~\cite{rublee2011orb}, BRISK~\cite{leutenegger2011brisk}, FREAK~\cite{alahi2012freak}, LATCH~\cite{levi2016latch},
BinBoost~\cite{trzcinski2015learning}.

On average, it takes 0.21 ms in a desktop CPU and 0.64 in a smartphone for the most accurate BEBLID implementation, with 512 bits, to process a $800\times640$ image with 2000 keypoints. This is roughly $20\times$ faster than LATCH, the most recent competing binary descriptor. The 256 bits implementation is slightly less accurate, but roughly $2\times$ faster. This means that our implementation of BEBLID with 256 bits is $4\times$ faster than OpenCV's ORB, the fastest binary descriptor in the literature, and $50\times$ faster than BinBoost, the best binary descriptor in terms of accuracy. Compared with other competing floating point approaches, BEBLID 512b is as fast as the 512f version of BELID-U-ADA and more than $20\times$ faster than BELID 512f, the comparable floating point descriptors. BEBLID 256b is roughly two orders of magnitude faster than SIFT.

The key for the computational efficiency of BELID and BEBLID lies in the use of very efficient WLs based on pairwise comparisons computed on the integral image. For this reason BEBLID computational requirements should be similar to those of ORB. The differences are caused by the fact that we extract our features in parallel, whereas the present ORB implementation in OpenCV does not. BELID is less efficient than BEBLID because it requires an extra multiplication of the WLs measurements, $\vh(\vw)$, with matrix $\mB$. However, BEBLID is as efficient as BELID-U-ADA, since in it $\mB$ is the identity.

From the results in this section we can conclude that BEBLID is the most efficient binary descriptor in the literature. Our new binary descriptor is the best compromise between mAP and speed. These results support the claim that our descriptor is a faster alternative to SIFT that is able to run in real-time on low performance devices. 

\begin{table}
\centering
\caption{Average description time per image, in milliseconds, of various descriptors in three platforms, two of them power-limited (Exynox Octa S and Snapdragon 855). The column "Size" reports the descriptor size in floating-point (f) or binary (b) values.}
\label{tab:execution_times}
\footnotesize
\begin{tabular}{c||c|c|c|c}
         & \multirow{2}{*}{Size}     & Intel Core i7 & Exynox & Snapdragon \\
         &                           & 8750H         & Octa   S& 855 \\ \hline
SIFT     & 128f            & 14.29   & 152.30 &  53.34 \\ \hline
ORB      & 256b            & 0.45    & 5.49   & 1.22   \\ \hline
BRISK    & 512b            & 0.92    & 8.27   & 1.92   \\ \hline
FREAK    & 512b            & 0.47    & 4.70   & 1.25   \\ \hline
LATCH    & 512b            & 5.21    & 62.78  & 8.33   \\ \hline
BinBoost$_{32}$-256 & 256b & 6.55    & 52.63  & 12.82  \\ \hline
BELID    & 512f            & 5.46    & 40.70  & 13.95  \\ \hline
BELID    & 256f            & 2.83    & 21.46  & 7.26   \\ \hline
BELID-U-ADA & 512f         & 0.25    & 2.27   & 0.69   \\ \hline
BEBLID  & 512b             & 0.21    & 2.09   & 0.64   \\ \hline
BEBLID  & 256b             & \textbf{ 0.11 }  & \textbf{ 1.32 }  & \textbf{0.42}\\
\hline
\end{tabular}
\end{table}

\section{Conclusion}
\label{sec:conclusiones}

In this paper we introduce BEBLID, the best non CNN-based binary descriptor in the state of the art in terms of accuracy and the most efficient in terms of computational requirements. In our experiments we proved that it is faster than the popular OpenCV implementation of ORB, the fastest descriptor in the literature. This is due to the use of very efficient image features, based on gray value differences computed with the integral image. In terms of accuracy BEBLID is better than BinBoost, the best binary descriptor in the literature, and close to SIFT, the ``gold standard'' reference. This is due to the discriminative scheme used to select the image features and the possibility of learning the feature scale, represented by the feature box size. Furthermore, we provide different BEBLID descriptors trained with unbalanced data sets, to model the asymmetry in the matching and retrieval problems,  which significantly improves the evaluation results.  

As discussed in the introduction, feature matching is required in many other higher level computer vision tasks. In most of them it is a mid-level process often followed by model fitting, e.g. RANSAC. This robust fitting step fixes the errors occurred in the matching procedure. This is possibly one of the reasons why SIFT is still the most widely used descriptor. Although SIFT is not the best performing approach in terms of accuracy, it provides a reasonable trade-off between accuracy and computational requirements. In the context of real-time performance on computationally limited devices, BEBLID represents the best trade-off as it is faster than ORB with an accuracy close to that of SIFT.

\section*{Acknowledgments}
The authors thank the anonymous reviewers for their comments.
The following funding is gratefully acknowledged. Iago Su{\'a}rez, grant Doctorado Industrial  DI-16-08966; Jos\'e M. Buenaposada and Luis Baumela, Spanish MINECO project TIN2016-75982-C2-2-R.

\bibliographystyle{plain}
\bibliography{refs}

\begin{thebibliography}{10}

\bibitem{agarwal2009building}
Sameer Agarwal, Noah Snavely, Ian Simon, Steven~M Seitz, and Richard Szeliski.
\newblock Building {Rome} in a day.
\newblock In {\em Proc. of International Conference on Computer Vision}, pages
  72--79. IEEE, 2009.

\bibitem{alahi2012freak}
Alexandre Alahi, Raphael Ortiz, and Pierre Vandergheynst.
\newblock {FREAK}: Fast retina keypoint.
\newblock In {\em Proc. Conference on Computer Vision and Pattern Recognition},
  pages 510--517. IEEE, 2012.

\bibitem{balntas2017hpatches}
Vassileios Balntas, Karel Lenc, Andrea Vedaldi, and Krystian Mikolajczyk.
\newblock Hpatches: A benchmark and evaluation of handcrafted and learned local
  descriptors.
\newblock In {\em Proc. Conference on Computer Vision and Pattern Recognition},
  pages 5173--5182, 2017.

\bibitem{balntas2015bold}
Vassileios Balntas, Lilian Tang, and Krystian Mikolajczyk.
\newblock {BOLD} - {Binary} online learned descriptor for efficient image
  matching.
\newblock In {\em Proc. Conference on Computer Vision and Pattern Recognition},
  pages 2367--2375, June 2015.

\bibitem{bay2006surf}
Herbert Bay, Tinne Tuytelaars, and Luc Van~Gool.
\newblock {SURF}: Speeded up robust features.
\newblock In {\em Proc. European Conference on Computer Vision}, pages
  404--417. Springer, 2006.

\bibitem{bjorkman2014}
Mårten Björkman, Niklas Bergström, and Danica Kragic.
\newblock Detecting, segmenting and tracking unknown objects using multi-label
  {MRF} inference.
\newblock {\em Computer Vision and Image Understanding}, 118:111 -- 127, 2014.

\bibitem{calonder2010brief}
Michael Calonder, Vincent Lepetit, Christoph Strecha, and Pascal Fua.
\newblock {BRIEF}: Binary robust independent elementary features.
\newblock In {\em Proc. European Conference on Computer Vision}, pages
  778--792. Springer, 2010.

\bibitem{han2015matchnet}
Xufeng Han, Thomas Leung, Yangqing Jia, Rahul Sukthankar, and Alexander~C.
  Berg.
\newblock {MatchNet}: Unifying feature and metric learning for patch-based
  matching.
\newblock In {\em Proc. Conference on Computer Vision and Pattern Recognition},
  pages 3279--3286, June 2015.

\bibitem{he2018local}
Kun He, Yan Lu, and Stan Sclaroff.
\newblock Local descriptors optimized for average precision.
\newblock In {\em Proc. Conference on Computer Vision and Pattern Recognition},
  pages 596--605, 2018.

\bibitem{leutenegger2011brisk}
Stefan Leutenegger, Margarita Chli, and Roland Siegwart.
\newblock {BRISK}: Binary robust invariant scalable keypoints.
\newblock In {\em Proc. of International Conference on Computer Vision}, pages
  2548--2555. IEEE, 2011.

\bibitem{levi2016latch}
G.~{Levi} and T.~{Hassner}.
\newblock {LATCH}: Learned arrangements of three patch codes.
\newblock In {\em IEEE Winter Conference on Applications of Computer Vision
  (WACV)}, pages 1--9. IEEE, 2016.

\bibitem{lowe1999object}
David~G Lowe.
\newblock Object recognition from local scale-invariant features.
\newblock In {\em Proc. of International Conference on Computer Vision},
  volume~2, pages 1150--1157. IEEE, 1999.

\bibitem{lowe2004distinctive}
David~G Lowe.
\newblock Distinctive image features from scale-invariant keypoints.
\newblock {\em International Journal of Computer Vision}, 60(2):91--110, 2004.

\bibitem{matas2002}
J.~Matas, O.~Chum, M.~Urban, and T.~Pajdla.
\newblock Robust wide baseline stereo from maximally stable extremal regions.
\newblock In {\em Proc. British Machine Vision Conference}, pages 36.1--36.10,
  2002.

\bibitem{mikolajczyk2005performance}
Krystian Mikolajczyk and Cordelia Schmid.
\newblock A performance evaluation of local descriptors.
\newblock {\em IEEE Transactions on Pattern Analysis and Machine Intelligence},
  27(10):1615--1630, 2005.

\bibitem{mishchuk2017working}
Anastasiia Mishchuk, Dmytro Mishkin, Filip Radenovic, and Jiri Matas.
\newblock Working hard to know your neighbor's margins: Local descriptor
  learning loss.
\newblock In {\em Advances in Neural Information Processing Systems}, pages
  4826--4837, 2017.

\bibitem{mur2015}
R.~{Mur-Artal}, J.~M.~M. {Montiel}, and J.~D. {Tard\'os}.
\newblock {ORB-SLAM}: A versatile and accurate monocular {SLAM} system.
\newblock {\em IEEE Transactions on Robotics}, 31(5):1147--1163, Oct 2015.

\bibitem{Nister2006}
D.~{Nister} and H.~{Stewenius}.
\newblock Scalable recognition with a vocabulary tree.
\newblock In {\em Proc. Conference on Computer Vision and Pattern Recognition},
  volume~2, pages 2161--2168, June 2006.

\bibitem{pernici2014object}
Federico Pernici and Alberto Del~Bimbo.
\newblock Object tracking by oversampling local features.
\newblock {\em IEEE Transactions on Pattern Analysis and Machine Intelligence},
  36(12):2538--2551, 2014.

\bibitem{rosten2006machine}
Edward Rosten and Tom Drummond.
\newblock Machine learning for high-speed corner detection.
\newblock In {\em Proc. European Conference on Computer Vision}, pages
  430--443. Springer, 2006.

\bibitem{rublee2011orb}
E.~{Rublee}, V.~{Rabaud}, K.~{Konolige}, and G.~{Bradski}.
\newblock {ORB}: An efficient alternative to {SIFT} or {SURF}.
\newblock In {\em Proc. of International Conference on Computer Vision}, pages
  2564--2571, Nov 2011.

\bibitem{schonberger2016structure}
Johannes~L Schonberger and Jan-Michael Frahm.
\newblock Structure-from-motion revisited.
\newblock In {\em Proc. Conference on Computer Vision and Pattern Recognition},
  pages 4104--4113, 2016.

\bibitem{shakhnarovich2005learning}
Gregory Shakhnarovich.
\newblock {\em Learning Task-Specific Similarity}.
\newblock PhD thesis, Massachusetts Institute of Technology, 2005.

\bibitem{simonyan2014learning}
Karen Simonyan, Andrea Vedaldi, and Andrew Zisserman.
\newblock Learning local feature descriptors using convex optimisation.
\newblock {\em IEEE Transactions on Pattern Analysis and Machine Intelligence},
  36(8):1573--1585, 2014.

\bibitem{Suarez18}
I.~{Suarez}, E.~{Mu\~noz}, J.~M. {Buenaposada}, and L.~{Baumela}.
\newblock {FSG}: A statistical approach to line detection via fast segments
  grouping.
\newblock In {\em Proc. of Int. Conf. on Intell. Robots Systems}, pages
  97--102, Oct 2018.

\bibitem{suarez2019belid}
Iago Su{\'a}rez, Ghesn Sfeir, Jos{\'e}~M. Buenaposada, and Luis Baumela.
\newblock {BELID}: Boosted efficient local image descriptor.
\newblock In {\em Proc. of Iberian Conference on Pattern Recognition and Image
  Analysis}, pages 449--460, Cham, 2019. Springer International Publishing.

\bibitem{tian2017l2net}
Y.~{Tian}, B.~{Fan}, and F.~{Wu}.
\newblock {L2-Net}: Deep learning of discriminative patch descriptor in
  euclidean space.
\newblock In {\em Proc. Conference on Computer Vision and Pattern Recognition},
  pages 6128--6136, July 2017.

\bibitem{tola2008fast}
Engin Tola, Vincent Lepetit, and Pascal Fua.
\newblock A fast local descriptor for dense matching.
\newblock In {\em Proc. Conference on Computer Vision and Pattern Recognition},
  pages 1--8. IEEE, 2008.

\bibitem{trzcinski2015learning}
Tomasz Trzcinski, Mario Christoudias, and Vincent Lepetit.
\newblock Learning image descriptors with boosting.
\newblock {\em IEEE Transactions on Pattern Analysis and Machine Intelligence},
  37(3):597--610, 2015.

\bibitem{Balntas2016tfeat}
Daniel~Ponsa Vassileios~Balntas, Edgar~Riba and Krystian Mikolajczyk.
\newblock Learning local feature descriptors with triplets and shallow
  convolutional neural networks.
\newblock In {\em Proc. British Machine Vision Conference}, pages
  119.1--119.11, September 2016.

\bibitem{von2010lsd}
Rafael~Grompone Von~Gioi, Jeremie Jakubowicz, Jean-Michel Morel, and Gregory
  Randall.
\newblock {LSD}: A fast line segment detector with a false detection control.
\newblock {\em IEEE Transactions on Pattern Analysis and Machine Intelligence},
  32(4):722--732, 2010.

\bibitem{wei2018kernelized}
X.~{Wei}, Y.~{Zhang}, Y.~{Gong}, and N.~{Zheng}.
\newblock Kernelized subspace pooling for deep local descriptors.
\newblock In {\em Proc. Conference on Computer Vision and Pattern Recognition},
  pages 1867--1875, June 2018.

\bibitem{winder2007learning}
Simon~AJ Winder and Matthew Brown.
\newblock Learning local image descriptors.
\newblock In {\em Proc. Conference on Computer Vision and Pattern Recognition},
  pages 1--8. IEEE, 2007.

\bibitem{wohlhart2015learning}
Paul Wohlhart and Vincent Lepetit.
\newblock Learning descriptors for object recognition and {3D} pose estimation.
\newblock In {\em Proc. Conference on Computer Vision and Pattern Recognition},
  pages 3109--3118, 2015.

\bibitem{zhang2019softmargin}
Linguang Zhang and Szymon Rusinkiewicz.
\newblock Learning local descriptors with a {CDF-Based} dynamic soft margin.
\newblock In {\em Proc. of International Conference on Computer Vision},
  October 2019.

\end{thebibliography}

\end{document}